\documentclass{article}
\pdfoutput=1
\DeclareRobustCommand{\rchi}{{\mathpalette\irchi\relax}}
\newcommand{\irchi}[2]{\raisebox{\depth}{$#1\chi$}}
\usepackage{mathrsfs}
\usepackage{mathptmx}
\usepackage{microtype}
\usepackage{wrapfig}
\usepackage{graphicx}
\usepackage{subcaption}
\usepackage{booktabs}
\usepackage{amsmath}
\usepackage{amssymb}
\DeclareMathOperator{\EX}{\mathbb{E}}
\usepackage{hyperref}

\usepackage{neurips_2021}
\usepackage[utf8]{inputenc} 
\usepackage[T1]{fontenc}    
\usepackage{hyperref}       
\usepackage{url}            
\usepackage{booktabs}       
\usepackage{amsfonts}       
\usepackage{nicefrac}       
\usepackage{microtype}      
\usepackage{xcolor}         
\hypersetup{
    colorlinks=false,
    pdfborder={0 0 0},
}
\usepackage{enumitem}
\title{OMASGAN: Out-of-Distribution Minimum Anomaly Score GAN for Sample Generation on the Boundary}
\author{%
  David S.~Hippocampus\thanks{Use footnote for providing further information
    about author (webpage, alternative address)---\emph{not} for acknowledging
    funding agencies.} \\
  Department of Computer Science\\
  Cranberry-Lemon University\\
  Pittsburgh, PA 15213 \\
  \texttt{hippo@cs.cranberry-lemon.edu} \\
}
\begin{document}
\maketitle
\thispagestyle{plain}
\begin{abstract}
Generative models trained in an unsupervised manner may set high likelihood and low reconstruction loss to Out-of-Distribution (OoD) samples. This increases Type II errors and leads to missed anomalies, overall decreasing Anomaly Detection (AD) performance. In addition, AD models underperform due to the rarity of anomalies. To address these limitations, we propose the OoD Minimum Anomaly Score GAN (OMASGAN). OMASGAN generates, in a negative data augmentation manner, anomalous samples on the estimated distribution boundary. These samples are then used to refine an AD model, leading to more accurate estimation of the underlying data distribution including multimodal supports with disconnected modes. OMASGAN performs retraining by including the abnormal minimum-anomaly-score OoD samples generated on the distribution boundary in a self-supervised learning manner. For inference, for AD, we devise a discriminator which is trained with negative and positive samples either generated (negative or positive) or real (only positive). OMASGAN addresses the rarity of anomalies by generating strong and adversarial OoD samples on the distribution boundary using only normal class data, effectively addressing mode collapse. A key characteristic of our model is that it uses any f-divergence distribution metric in its variational representation, not requiring invertibility. OMASGAN does not use feature engineering and makes no assumptions about the data distribution. The evaluation of OMASGAN on image data using the leave-one-out methodology shows that it achieves an improvement of at least 0.24 and 0.07 points in AUROC on average on the MNIST and CIFAR-10 datasets, respectively, over other benchmark and state-of-the-art models for AD.
\end{abstract}
\section{Introduction} \label{sec:dsfas2321aassdaa21asa}
In spite of progress in Anomaly Detection (AD) ushered in by Generative Adversarial Networks (GAN) \cite{15}, models learn to assign high probability to the seen data but are not trained to assign zero probability to Out-of-Distribution (OoD) samples.
During inference, anomalies might be still assigned a non-zero probability, leading to false negatives \cite{28}.
To address such limitations, we propose the OoD Minimum Anomaly Score GAN (OMASGAN).
OMASGAN generates OoD anomalous samples close to the boundary having a minimum anomaly score around the data.
It performs (i) retraining by including the learned boundary, and (ii) AD with negative sampling and training \cite{42, 41}.
Our methodology can be used with other GAN formulations including \cite{44, 34}.
Our contributions are:
\begin{itemize}[leftmargin=*]
    \item We propose OMASGAN, a generalized methodology for AD to more accurately learn the underlying distribution of the data. OMASGAN performs retraining by including the boundary of the support of the data distribution which is generated by our negative data augmentation technique.
    \item To address the rarity of anomalies, we create abnormal samples using data only from the normal class and find the minimum-anomaly-score samples on the f-divergence boundary, without requiring likelihood and invertibility.
    OMASGAN avoids mode collapse and makes no assumptions about the data distribution.
    It does not use feature engineering and human intervention is not required.
    \item OMASGAN performs self-supervised learning to improve both unsupervised learning, i.e. learning the underlying data distribution which may be multimodal with a support with disjoint components, and AD by retraining including the generated anomalous samples on the distribution boundary.
    \item We train a discriminator to separate the data distribution from its complement and use it as an inference mechanism for AD.
    The evaluation of OMASGAN using the Leave-One-Out (LOO) methodology shows that it achieves state-of-the-art performance in the Area Under the Receiver Operating Characteristics curve (AUROC), outperforming benchmarks. 
    Our model is also evaluated using One-Class-Classification (OCC), outperforming GAN- and AE-based benchmark models.
\end{itemize}

\begin{figure*}[!tbp]
	\centering \includegraphics[width=0.791\textwidth]{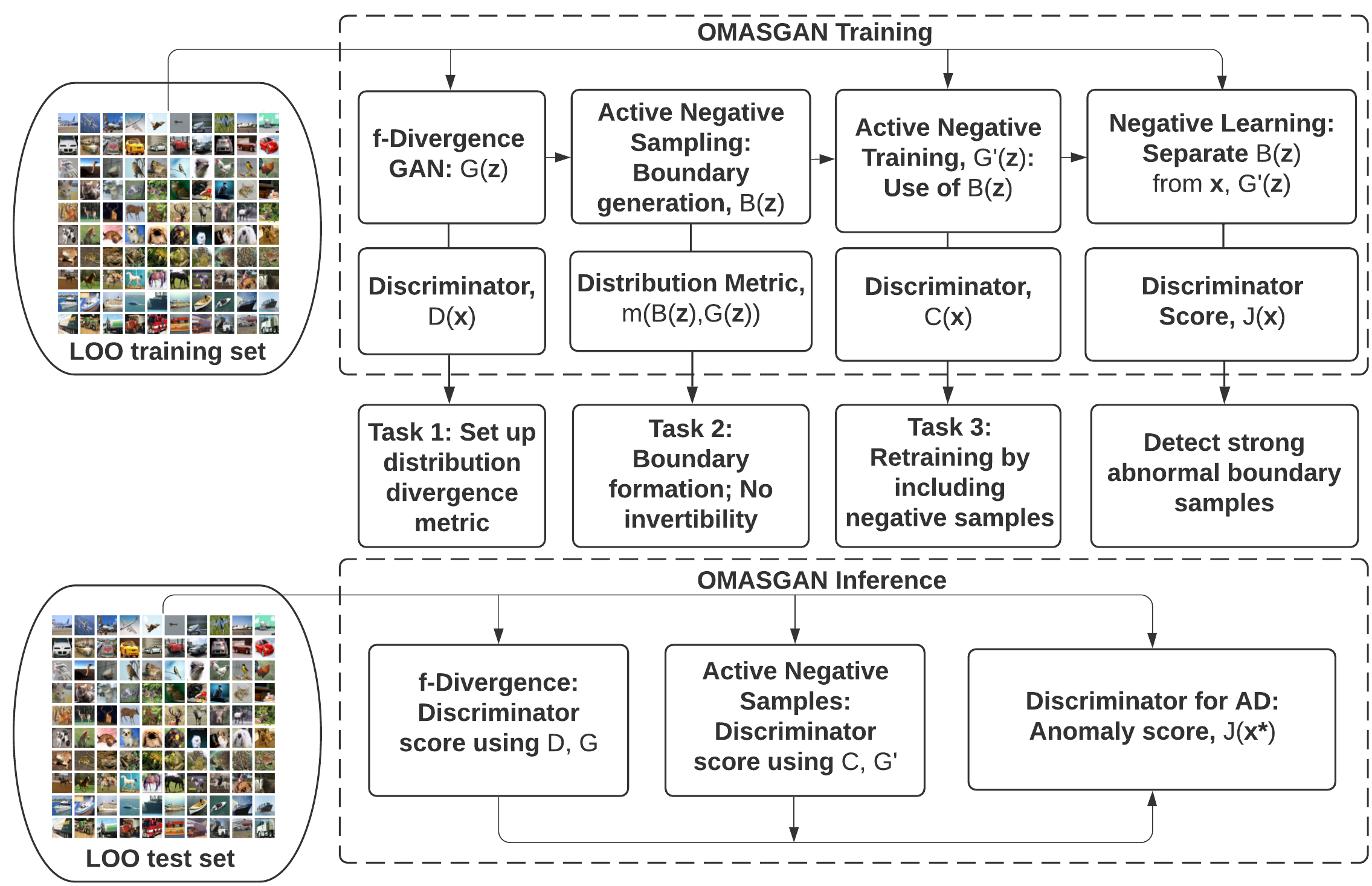}
	\caption{Flowchart of OMASGAN which generates minimum-anomaly-score OoD samples on the distribution boundary and subsequently uses the generated boundary to train a discriminator for AD.}
	\label{fig:1}
\end{figure*}

\section{Our Proposed OMASGAN Algorithm} \label{sec:sfhghjsf7safgs} \label{sec:asdfsadfsdafsfsdfss}
\textbf{Overview.} We propose OMASGAN to solve the problem of models setting high likelihood and low reconstruction loss to OoD samples which leads to Type II errors (misses of anomalies), decreasing AD performance \cite{28}.
Figure~\ref{fig:1} shows the flowchart of OMASGAN:
(a) Train a f-divergence-based GAN to obtain the generator, $G(\textbf{z})$, where $\textbf{z}$ is a latent variable.
We denote the latent space by $\mathscr{Z} \in \mathbb{R}^l$.
The GAN samples, $G(\textbf{z})$, and the data, $\textbf{x}$, lie in the data space, $\rchi \in \mathbb{R}^k$, where $l < k$.
(b) Train a boundary data generator, $B(\textbf{z})$, to obtain boundary samples to be used as negative points, for active negative sampling.
We compute the statistical divergence between $B$ and $G$, and we find the boundary of $G$ using any f-divergence in its variational representation.
The $B(\textbf{z})$ boundary samples lie in $\rchi$.

OMASGAN generates samples corresponding to a generalized notion of the boundary of the support of the data distribution, which is the set of points such that they are OoD and have a minimum anomaly score measured as the f-divergence.
OMASGAN generates OoD samples and we incorporate these OoD Minimum Anomaly Score (OMAS) samples in our algorithm.
(c) Perform negative retraining using the OMAS samples and the implicit distributions from (a) and (b), and train the generator $G'(\textbf{z})$ using a discriminator, $C(\textbf{x})$.
The samples $G'(\textbf{z})$ lie in $\rchi$.
OMASGAN performs model retraining and subsequently trains a discriminator for AD using the learned distribution boundary.
We train the discriminator $J(\textbf{x})$ for AD using the OMAS samples and active negative training.
During inference, for AD, we compute our proposed anomaly score and detect abnormal samples using $J$ and $G'$.

OMASGAN comprises the optimization tasks:
\textbf{Task 1. Establishing a distribution metric.}
We train a f-divergence-based GAN model to learn the data, $\textbf{x}$. Using $\textbf{z} \sim p_{\textbf{z}}$, $\textbf{x} \sim p_{\textbf{x}}$, and $G(\textbf{z}) \sim p_g$,
\begin{align}
\begin{split}
\text{arg} \, \, \text{min}_G \, \, \text{max}_D \, \, \EX_{\textbf{x}} \log(D(\textbf{x})) + \EX_{\textbf{z}} \log(1 - D(G(\textbf{z})))\text{.}
\end{split}
\end{align}
Using conjugate functions, $f^*$, \cite{34, 44} and for example $D(\textbf{x})=1/(1+\exp(-V_D(\textbf{x}))) \, \text{and} \, g_f(\textbf{v})=-\log(1+\exp(-v))$, the f-divergence optimization objective of OMASGAN Task 1 is given by
\begin{align}
\begin{split}
\text{arg} \, \, \text{min}_G \, \, \text{max}_D \, \, \EX_{\textbf{x}} \, g_f(V_D(\textbf{x})) \, - \, \EX_{\textbf{z}} \, f^*(g_f(V_D(G(\textbf{z})))) \text{.}
\end{split}
\end{align}

\textbf{Task 2. Formation of the data distribution's boundary.} To perform active negative sampling, we train the distribution boundary model, $B(\textbf{z})$. The parameters~of~$B(\textbf{z})$~are~$\pmb{\theta_b}$. The optimization is
\begin{align}
\begin{split}
\text{arg} \ \, \text{min}_{\pmb{\theta_b}} \, & - m( B(\textbf{z}; \pmb{\theta_b}), G(\textbf{z})) + \mu \ d( B(\textbf{z}; \pmb{\theta_b}), G(\textbf{z}) ) + \nu \ s( B(\textbf{z}; \pmb{\theta_b}), \textbf{z} ) \label{eq:qwerqwrq}\\
\end{split}
\end{align}
where $m(B, G)$ is the distribution metric from Task 1, i.e. any f-divergence in its variational representation expressed in terms of the conjugate function, $f^*(t)$, as in (7) in \cite{34}, where $t$ is a variational function taking as input a sample and returning a scalar.
The special cases of KL and Pearson are $f^*(t)=\text{exp}(t-1)$ and $f^*(t)=0.25t^2+t$, respectively.
The first term in \eqref{eq:qwerqwrq} is a \textit{strictly decreasing function} of a distribution metric.
This divergence is between the boundary samples and the data where $m( B(\textbf{z}), G(\textbf{z}))$ is the distribution metric of choice, such as Jensen-Shannon of the original GAN, Kullback–Leibler (KL), and Pearson Chi-Squared of the Least Squares GAN \cite{27} to address saturation and mode collapse.
The proposed boundary model does not need probability and invertibility obviating the rarity sampling complexity problem by using a decreasing function of a statistical divergence distribution metric, including any f-divergence, to find the data distribution boundary.

\textit{Generating the OMAS distribution and OMAS samples}: \label{sec:secasdf232a}
We find minimum-anomaly-score OoD samples and perform learned negative data augmentation eliminating the need for feature extraction and human intervention.
This strengthens \textit{applicability}.
We use a f-divergence GAN discriminator to compute the distribution metric, $m( B, G )$.
The first two terms in \eqref{eq:qwerqwrq} lead the $B(\textbf{z})$ samples to the boundary of $p_g$.
We compute the boundary with $l_p$-norm distance and dispersion regularization.
We denote the $l_p$-norm distance between the point $B(\textbf{z})$ and the set $\textbf{x}$ by $d( B(\textbf{z}), \textbf{x} )$.
To capture all the modes of the data distribution, avoid mode collapse \cite{11, 2}, and generate OMAS samples, we use the scattering measure $s( B(\textbf{z}), \textbf{z} )$.
The distance measure, $d( B(\textbf{z}), G(\textbf{z}) )$, and $s( B(\textbf{z}), \textbf{z} )$ are given by
\begin{align}
& d( B(\textbf{z}_i), G(\textbf{z}) ) \, = \, \text{min}_{ j = 1, \dots, Q} \, || B(\textbf{z}_i) - G(\textbf{z}_{j}) ||_2 \label{eq:qwsfsdfgserasdfdasdfqafgsgswrq} \\
& s( B(\textbf{z}_i), \textbf{z}_i ) \, = \, \frac{1}{N-1} \sum_{j=1, \, j \neq i}^N \dfrac{ || \textbf{z}_i - \textbf{z}_j||_2 }{ || B(\textbf{z}_i) - B(\textbf{z}_j) ||_2 } \label{eq:qwsfgserqafgsgswrq}
\end{align}
where the batch and inference sizes are denoted by $N$ and $Q$, respectively.
A variation of our point-set distance, $d( B(\textbf{z}), G(\textbf{z}) )$, can be obtained by the Chamfer statistical divergence \cite{32}.
Good modeling is achieved by finding the boundary of a model of $\textbf{x}$ rather than of $\textbf{x}$.
OMASGAN finds the minimum-anomaly score OoD samples, performs active sampling of negative examples, and generates strong and specifically adversarial anomalies.
Strong anomalies lie close to the data distribution boundary while \textit{adversarial anomalies} are strong anomalies that are close to high-probability data samples.

\begin{figure*}[!tbp]
	\centering \includegraphics[width=0.785\textwidth]{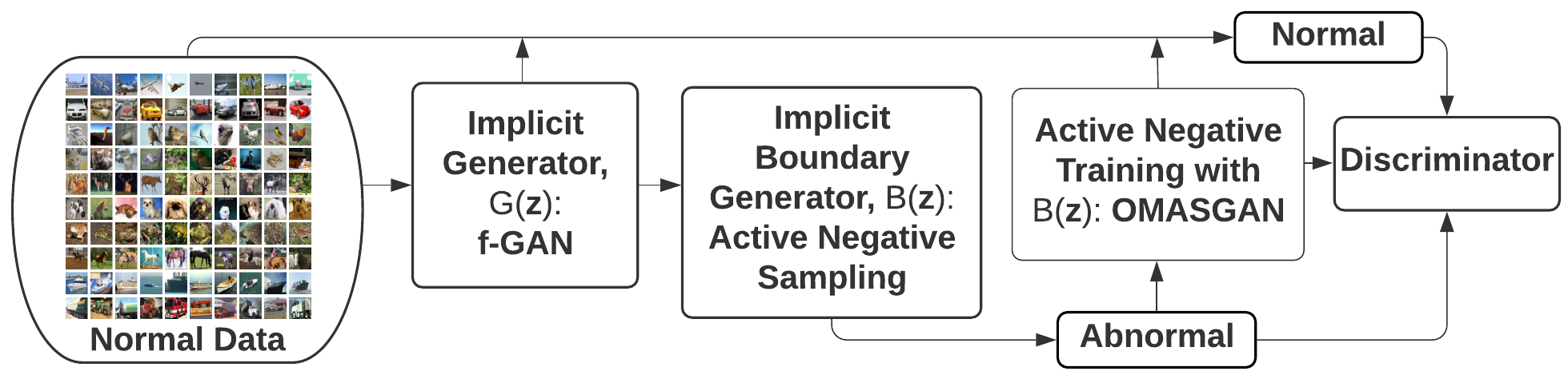}
	\caption{Training of OMASGAN using negative sampling, by generating \textit{strong abnormal} samples.}
	\label{fig:2}
\end{figure*}

We use the output of Task 1, $G(\textbf{z})$, as it has established and set up a distribution metric in $\rchi$.
In Task 1, OMASGAN uses a f-divergence-based GAN for data distribution generation. Task 2 performs active negative \textit{sampling} and trains a boundary generator to form the boundary.
We avoid mode collapse and use a relaxed form of the constrained optimization problem of maximizing dispersion subject to being \textit{on the f-divergence boundary}.
In \eqref{eq:qwerqwrq}, $\mu$ and $\nu$ are hyperparameters.
In Task 3, OMASGAN performs active negative training with the minimum-anomaly-score OoD boundary samples, and trains a discriminator, $J(\textbf{x})$, to classify normal data and abnormal samples, as shown in Figure~\ref{fig:2}.

\textbf{Task 3. Active negative training.} 
\textit{Separation of generated and real normal from generated abnormal data for AD:}
To address the learning-OoD-samples problem of $G$ \cite{28, 21}, we perform retraining for AD by including the negative samples generated by our negative samples augmentation methodology in Task 2.
OMASGAN introduces self-supervision by using the learned data distribution boundary from Task 2.
We \textit{negatively retrain} by including the abnormal OoD $B(\textbf{z})$.
To train $G’(\textbf{z})$, we use
\begin{align}
\begin{split}
\text{arg} \, \ \text{min}_{G’} \, \ \text{max}_C \ & \, \EX_{\textbf{z}} \log(1 - C(G’(\textbf{z}))) \, + \, \alpha \EX_{\textbf{x}} \log(C(\textbf{x})) \label{eq:dasfadfadsfa1xsx1x} \\
& + \, \beta \EX_{\textbf{z}} \log(1-C(B(\textbf{z}))) \, + \, \gamma \EX_{\textbf{z}} \log(C(G(\textbf{z})))
\end{split}
\end{align}
where $G'(\textbf{z}) \sim p_{g'}$ lie in $\rchi$ and where $C$ is a discriminator which computes distribution metrics and f-divergences, as in \cite{51, 6}.
To calculate statistical divergences between distributions, and in this case between $B(\textbf{z})$ and $(\textbf{x}, G(\textbf{z}))$, we use a discriminator and a \textit{weighted sum of f-divergences} and probability metrics \cite{51, 6}.
The proposed nested optimization in \eqref{eq:dasfadfadsfa1xsx1x} comprises four terms and outputs the learned mappings $C: \rchi \rightarrow \mathbb{R}$, where the data space is denoted by $\rchi$, and $G': \mathscr{Z} \rightarrow \rchi$, where $\mathscr{Z}$ is the latent space.
The trainable models are the implicit generator $G'(\textbf{z})$ and the discriminator $C(\textbf{x})$, while $B$ and $G$ are non-trainable.
The first and fourth terms of the minmax optimization force the generated samples to the data, as in Rumi-GAN \cite{6}.
The third term forces the generated samples away from our generated strong anomalies, which are near the support boundary of the data distribution and close to high-probability data samples.
The discriminator, $C(\textbf{x})$, is trained to separate $B(\textbf{z})$ from $(\textbf{x}, G(\textbf{z}))$.
$G'$ learns the data, $\textbf{x}$, keeping away from and avoiding the generated abnormal $B(\textbf{z})$.

OMASGAN computes the optimal points for negative sampling in Task 2.
It performs active negative training using the boundary.
The samples from $B(\textbf{z})$ are \textit{the closest points} to the normal class data and optimal retraining is performed.
This leads to improvements with respect to \cite{51, 35, 41}.
We bias the GAN generator to avoid negative samples, improving the AD performance.
In contrast to \cite{13, 8}, we do not use feature engineering; this strengthens scalability.
OMASGAN performs \textit{automatic negative samples augmentation} and retraining for AD by eliminating the need for feature extraction, human intervention, and ad hoc methods. This strengthens applicability and generalization.
Our learned negative data augmentation and retraining methodology incorporates prior knowledge through OoD boundary samples. It is complementary to traditional in-distribution data augmentation methods.

\textbf{Detection of strong abnormal boundary samples.}
To address the learning-OoD-samples problem and to perform active negative training for OoD detection, we train the discriminative model $J(\textbf{x})$,
\begin{align}
\begin{split}
\text{arg} \ \, \text{max}_J \ \, & \EX_{\textbf{z}} \log(J(B(\textbf{z}))) + \delta \, \EX_{\textbf{x}} \log(1 - J(\textbf{x})) + (1-\delta) \EX_{\textbf{z}} \log(1 - J(G’(\textbf{z})))\text{.} \label{eq:sdafasdf98sadf766sdafas}
\end{split}
\end{align}
\textit{Inputs to final negative training}: 
The inputs to $J$ are $\textbf{x}$ and samples from $\rchi$.
The inputs to $G'(\textbf{z})$ and $B(\textbf{z})$ are samples from $\mathscr{Z}$.
\textit{Output}: $J$ which learns to separate $B(\textbf{z})$ from normality, $\textbf{x}$ and $G'(\textbf{z})$.

\textbf{Our inference mechanism.}
We use the Anomaly Discriminator, $J$, and the f-divergence distribution metric for AD.
The f-divergence is used for training and we also use it during \textit{inference}.
The GAN discriminator computes f-divergences; for the distributions $P$ and $R$, we write this metric as fD($P$, $R$).
We compute fD($G'$, $\delta_\textbf{x}^*$) for a test sample $\textbf{x}^*$ where $G'$ is the learned data distribution after retraining and $\delta_\textbf{x}^*$ is a Dirac function centered at $\textbf{x}^*$.
For \textit{any} $\textbf{x}^* \in \rchi$, the Anomaly Score (AS) is given by
\begin{align}
\begin{split}
AS(\textbf{x}^*) \, = \, J(\textbf{x}^*) \, + \, \lambda \, \text{fD}(G', \delta_\textbf{x}^*)\text{.}
\end{split}
\end{align}
The classification decision is: $\textbf{x}^*$ is from the normal class if $J(\textbf{x}^*)+\lambda \, \text{fD}(G',\delta_\textbf{x}^*) < \tau$, where $\tau$ is a predefined threshold, and $\textbf{x}^*$ is abnormal otherwise.
By including negative samples, $J(\textbf{x})$ learns to discriminate between the data distribution and \textit{its complement}.
We leverage $J(\textbf{x})$ to detect OoD samples \cite{51}.
OMASGAN generates minimum-anomaly-score OoD samples and subsequently trains $J$ using the learned/generated boundary.
We use negative training by generating OoD data on the data distribution's boundary since our $B(\textbf{z})$ model samples from the boundary of the data distribution.

\textbf{Properties of OMASGAN.} \label{sec:sdfsa2342adadfd234a}
{{\color{black} \textit{Task 1:}
Let $\textbf{x} \sim p_{\textbf{x}}$ lie in $\rchi$, $\textbf{z} \sim p_{\textbf{z}}$ in $\mathscr{Z}$ and $G(\textbf{z}) \sim p_g$ in $\rchi$.
Task 1 attains a global optimum, $p_g = p_{\textbf{x}}$, and convergence to a local optimum is guaranteed \cite{34}.
\textit{Task 2 Global Properties:}
Let $L(\pmb{\theta_b}, \textbf{z}, B, G)$ be the loss in \eqref{eq:qwerqwrq} which is a continuous function of $\pmb{\theta_b}$. Let the set over which $\pmb{\theta_b}$ vary be compact. Then, \eqref{eq:qwerqwrq} attains a global minimum at $\pmb{\theta_b}^*$.
A continuous function defined on a compact set attains a global minimum and maximum: Weierstrass Extreme Value Theorem. \eqref{eq:qwerqwrq} attains a global minimum; it is continuous as a function of the parameters of $B$ and $G$, \eqref{eq:qwerqwrq}-\eqref{eq:qwsfgserqafgsgswrq} are continuous, and the set over which $\pmb{\theta_b}$ vary is compact, i.e. \textit{sufficient condition}.

\textbf{Local properties.} \label{sec:sadfdfe3a}
\textit{Proposition:} Let $\pmb{\tilde{\theta_b}}$ be the locality where our Task 2 converges. Then, wherever we terminate, that point leads $B(\textbf{z}; \pmb{\tilde{\theta_b}})$ to be a distribution on the boundary.
We find $\pmb{\theta_b}$ such that the distribution on $\rchi$ is evenly distributed on the boundary surface manifold, defined as the set of parameters $A^* \triangleq \text{arg} \, \text{min} \, \{ -m(c) + \mu \, d(c) \, | \, c \}$. \textit{Necessary condition:} $\nabla (-m(c) + \mu  \, d(c)) = 0$ which defines the tangent plane to the manifold.
We find a boundary distribution; this distribution is evenly distributed on the support boundary of the data distribution.
Solving $A^{**} = \text{arg} \, \text{min} \, \{ s(c) \, | \, c \in A^* \}$ is a hard constrained optimization problem.
We introduce a \textit{regularized loss} instead of a much harder constrained optimization.
We regularize and solve $\text{arg} \, \text{min} \, -m(c) + \mu  d(c) + \nu  s(c)$.
By choosing the hyperparameters $\mu$ and $\nu$ appropriately, we find points in $A^{**}$.
Gradient Descent finds a local minimum of this combined loss.
Using Gradient Descent \cite{53}, the solution to the first two terms in \eqref{eq:qwerqwrq} and \eqref{eq:qwsfsdfgserasdfdasdfqafgsgswrq}, with the third term in \eqref{eq:qwsfgserqafgsgswrq}, lies on the OMAS manifold.
If the loss is locally convex, \eqref{eq:qwerqwrq} attains a local minimum.
When Task 2 converges at $\pmb{\tilde{\theta_b}}$, dispersion is maximized.
Mode collapse is eliminated because of the scattering measure in \eqref{eq:qwerqwrq} and its combination with the other loss terms.

\textbf{Global and local properties of Task 3.} \label{sec:hdsgahg32asf1d1}
Let $p_{b}$ be the boundary distribution from \eqref{eq:qwerqwrq}. Task 3 in \eqref{eq:dasfadfadsfa1xsx1x} attains a global optimum, $p_{g'} = p_{\textbf{x}}$. Using both positive and negative samples, $G'$ learns the distribution of the samples from the positive class \cite{6, 51}. The global optimum at $p_{g'} = (1+\beta) p_{\textbf{x}} - \beta \, p_{b}$ subject to both $\beta \geq \alpha + \gamma - 1$ and $\alpha + \gamma \in [0, 1]$ subsumes that at $p_{g'} = p_{\textbf{x}}$ as a special case and using alternating Stochastic Gradient Descent, convergence to a local optimum is guaranteed \cite{6}.

Let $L_J(\pmb{\theta_j}, \textbf{z}, \textbf{x}, J, B, G')$ be the \textit{loss} in \eqref{eq:sdafasdf98sadf766sdafas}. The set over which $\pmb{\theta_j}$ vary is compact and \eqref{eq:sdafasdf98sadf766sdafas} is continuous as a function of the model parameters. It attains a global \textit{minimum} at $\pmb{\theta_j}^*$ such that $L_J(\pmb{\theta_j}^*, \textbf{z}, \textbf{x}, J, B, G')$ is \textit{lowest}, using the Extreme Value Theorem (\textit{sufficient condition}).
\textit{Local properties:}
Let $\pmb{\tilde{\theta_j}}$ be the point where \eqref{eq:sdafasdf98sadf766sdafas} converges. Wherever OMASGAN terminates, that $\pmb{\tilde{\theta_j}}$ leads $J(\textbf{x}; \pmb{\tilde{\theta_j}})$ to separate the data distribution from its complement. The $B(\textbf{z})$ model samples points from the distribution boundary as $B(\textbf{z})$ is an implicit distribution on the boundary producing \textit{OMAS points for negative training}.

\section{Related Work} \label{sec:dsafas23422ada1a1a}
OMASGAN addresses the \textit{rarity of abnormal data} and provides negative data augmentation by creating strong OoD data on the distribution boundary, unlike \cite{46, 42}. Our methodology performs sampling of negative points, creates optimal points for negative training \textit{closest to the data}, and does not need to know any data features, in contrast to \cite{41} which needs to have a process that programmatically creates the OoD examples using image transformations.
OMASGAN \textit{learns to generate} OoD samples.
We perform retraining using active negative sampling setting the boundary points as \textit{strong anomalies}.
This is our contribution and this differs from creating OoD samples by using (i) low-epoch reconstructions \cite{51, 35}, (ii) rotated features \cite{41}, and (iii) a CVAE \cite{9}.
Old is Gold (OGNet) uses weak anomalies far from the boundary, low-quality reconstructions, and pseudo-anomalies generated in an \textit{ad hoc manner} without any guarantee of coverage of the OoD part of the data space.
OGNet uses a pseudo-anomaly module to create OoD points.
It uses a \textit{restrictive definition of anomaly} as single-epoch blurry reconstructions.
It modifies the role of the discriminator (f-divergence distribution metrics) by using an Autoencoder (AE) to distinguish good from bad quality reconstructions.
For good and poor quality samples, a generator and an old state of the same AE generator are used.
Anomalies that are far away from the boundary are also created by \cite{9}.

Minimum Likelihood GAN (MinLGAN) and FenceGAN generate boundary samples to subsequently use the discriminator score for AD \cite{49, 30}.
In contrast to the Boundary of Distribution Support Generator (BDSG) \cite{11}, OMASGAN uses any f-divergence distribution metric, no invertibility, and a discriminator for AD.
OMASGAN maximizes dispersion subject to being on the f-divergence boundary without mode collapse using a \textit{relaxed form} of a constrained optimization problem.
Our self-supervised learning methodology involves model retraining by including the learned distribution boundary.
As opposed to others, OMASGAN makes no assumptions about the data distribution and is not ad hoc.
AD is an automatic outcome of our generalized OMASGAN methodology which can improve upon many of the existing techniques to more accurately and more robustly learn $p_{\textbf{x}}$.

The \textit{rarity of anomalies} is not addressed by \cite{37, 6} which highlight the benefit of supervision.
Rarity is addressed by GEOM \cite{13}, GOAD \cite{8}, and Deep Robust One Class Classification (DROCC) \cite{16}.
GEOM trains a multi-class model to discriminate between geometric image transformations, horizontal flipping, translations, and rotations.
It learns feature detectors that identify anomalies based on the model's softmax activation statistics.
The classification-based model GOAD generalizes transformation-based methods using affine and geometric transformations.
DeepSVDD \cite{36} minimizes the volume of a hypersphere to enclose the data in the latent space using a deep kernel-based AD loss.
Mappings of anomalies lie outside, while mappings of normality lie within the hypersphere.
However, DeepSVDD suffers from \textit{representation collapse}.
Representations richer than a hypersphere are needed.
DROCC is robust to representation collapse and assumes that the data lie on a locally linear well-sampled low-dimensional manifold, but it does not find the distribution boundary.

Table~\ref{tab:1} presents a summary of the main characteristics of OMASGAN and the benchmarks.
We focus on the key properties of OMASGAN and on the different approaches of GAN, AE, and classification.
We highlight the differences in the losses (active negative training, boundary loss) and in the inference mechanism (discriminator anomaly score).
In contrast to others, OMASGAN performs active negative sampling and introduces self-generated labels and supervision. It is a GAN and this is desirable since (a) a distribution metric is established in $\rchi$, (b) GANs and \textit{distribution metrics} have been shown to outperform AEs and distance metrics, and (c) contextual anomalies that have shared features with the data cannot be detected using a reconstruction anomaly score, i.e. near low probability samples.

\begin{table*}[!tbp]
\caption{ \centering Properties and architecture characteristics of OMASGAN and benchmarks for AD.}
\label{tab:1}
\vskip -0.0706in
\vskip -0.0706in
\vskip 0in
\begin{flushleft}
\begin{scriptsize}
\begin{sc}
\begin{tabular} 
{p{1.4cm} p{0.6cm} p{0.4cm} p{0.4cm} p{0.4cm} p{0.4cm} p{0.4cm} p{0.5cm} p{0.4cm} p{0.4cm} p{0.4cm} p{0.5cm} p{0.6cm} p{0.25cm} p{0.4cm}}
\toprule
\midrule
       & OMAS & MinL & Fence & EGB & Ano & Tail & BDSG & OG & Deep & VAE & GAN & GEOM & GO & DR\\
       & GAN & GAN & GAN & AD & GAN & GAN & & Net & SAD & & omaly & & AD & OCC\\
\midrule
&   & \cite{49} & \cite{30} & \cite{52} & \cite{39} & \cite{12} & \cite{11} & \cite{51} & \cite{37} & \cite{52} & \cite{4} & \cite{13} & \cite{8} & \cite{16}\\
\midrule
\midrule
GAN-based & 
\hspace{3pt} $\surd$ & \hspace{3pt} $\surd$ & \hspace{3pt}$\surd$ & \hspace{3pt}$\surd$ & \hspace{2pt}$\surd$ & \hspace{2pt}$\surd$  &      &      &    & & &  & &    \\
\midrule
AE-based &  &   &  &  &  & & & \hspace{3pt}$\surd$ & \hspace{5pt}$\surd$ & $\hspace{5pt}\surd$  & $\hspace{5pt}\surd$   &     &     &     \\
\midrule
Classification &  &   &  &  &  &  & 
&  &  &  &  & \hspace{5pt}$\surd$  & \hspace{1pt}$\surd$ &
\\
\midrule
\midrule
ANST & \hspace{3pt} $\surd$ & \hspace{5pt}$\surd$  
&  &  &  &  &  & \hspace{5pt}$\surd$  &  
& &  & \hspace{5pt}$\surd$  & \hspace{1pt}$\surd$  &  \hspace{5pt}$\surd$   \\
\midrule
NT & \hspace{3pt} $\surd$ & \hspace{3pt}$\surd$  &  &  &  & & & \hspace{5pt}$\surd$ & \hspace{5pt}$\surd$  &   &  & \hspace{5pt}$\surd$ & \hspace{1pt}$\surd$ &  \hspace{5pt}$\surd$\\ 
\midrule
Boundary & \hspace{3pt} $\surd$ &   & \hspace{3pt}$\surd$ &  & & \hspace{1pt}$\surd$  & \hspace{4pt} $\surd$  &  &  &  &  &  &  &  \\
\midrule
fD & \hspace{5pt}$\surd$ &   &  &  &  &  &  &  &  &  & & & & \\
\midrule
Discr. AS & \hspace{4pt} $\surd$ & \hspace{4pt}$\surd$ & \hspace{4pt}$\surd$ & \hspace{4pt}$\surd$ & \hspace{2pt}$\surd$ &  &  & \hspace{4pt}$\surd$
 &  &  &  &  &  & \hspace{3pt}$\surd$ \\
\midrule
Reconstruction &   &    &   &  \hspace{4pt}$\surd$ &  \hspace{2pt}$\surd$ &  &  \hspace{4pt}$\surd$  &  \hspace{4pt}$\surd$ &  \hspace{4pt}$\surd$ & \hspace{4pt}$\surd$ & \hspace{4pt}$\surd$   &    &  &  \hspace{4pt}$\surd$  \\
\midrule
Likelihood &  & \hspace{2pt}$\surd$ & &  &  & \hspace{2pt}$\surd$ & \hspace{4pt}$\surd$  &  &  & \hspace{4pt}$\surd$ & &  &  & \\
\midrule
\bottomrule
\end{tabular}
\end{sc}
\\
\vskip 0.05in
ANST = Active Negative Sampling and Training; NT = Negative Training; fD = f-Divergence; Discr. AS = Discriminator Anomaly Score\\
\end{scriptsize}
\end{flushleft}
\vskip -0.1in
\end{table*}

\section{Evaluation of OMASGAN} \label{sec:gasdfjg6asdfa}
We evaluate OMASGAN using the AUROC.
The LOO methodology is used; we set $K$ classes of a dataset with $(K + 1)$ classes as the normal class and the \textit{leave-out class} as the abnormal class.
LOO evaluation is more challenging than One Class Classification (OCC) evaluation used by MinLGAN, OGNet, and \cite{43, 31, 20} which is setting a single class of a dataset as normal and all the remaining classes of the dataset as the abnormal class.
The LOO evaluation methodology is more realistic and closer to real-world scenarios than OCC evaluation \cite{3}, as OCC oversimplifies the AD problem.

\textbf{Models.} We train dense fully-connected network architectures for synthetic data and Convolutional Neural Networks (CNN) with batch normalization for image data. We use the recently developed f-divergence-based KL-Wasserstein GAN model (KLWGAN) \cite{44} and the f-GAN model \cite{34}.

\textbf{Benchmarks.}
We evaluate OMASGAN using LOO on MNIST and compare it to the GAN models EGBAD \cite{52} and AnoGAN \cite{39}, to the likelihood-based AD models BDSG and TailGAN \cite{11, 12}, and to the AE models GANomaly and VAE.
We also evaluate OMASGAN using LOO on CIFAR-10 and compare it to EGBAD, AnoGAN, BDSG, GANomaly, VAE, and FenceGAN \cite{30}.
We also compare OMASGAN to GOAD, GEOM, DROCC, and DeepSVDD using OCC on CIFAR-10.

\textbf{Implementation and training.}
In the LOO setting, the MNIST and CIFAR-10 datasets are transformed into ten AD tasks, where one class is used as the abnormal class and the remaining classes are treated as samples from the normal class.
In OCC, CIFAR-10 is transformed into ten AD tasks, where one class is the normal class and the remaining classes are abnormal.
The training and validation sets for OMASGAN contain only samples from the normal class.
This scenario resembles typical real-world situations where anomalies are rare and not available during training \cite{31, 45}.
We traverse a set of hyperparameters empirically chosen for Task~2 in (3) and Task~3 in (6) and (7) using a validation set that contains only normal class samples.
The validation set of the normal class samples is also used to early stop the training.
We select the hyperparameters that lead to the lowest values for the individual loss terms.
We observe that from task to task, as well as from dataset to dataset, each term of our losses is not sensitive to the hyperparameters.
The hyperparameter values that achieve good performance do not vary across AD tasks, and they prove to be robust across datasets too.

\begin{wrapfigure}{r}{0.49\textwidth} 
    \vspace{-7pt}
    \centering
    \includegraphics[height=0.379\textwidth]{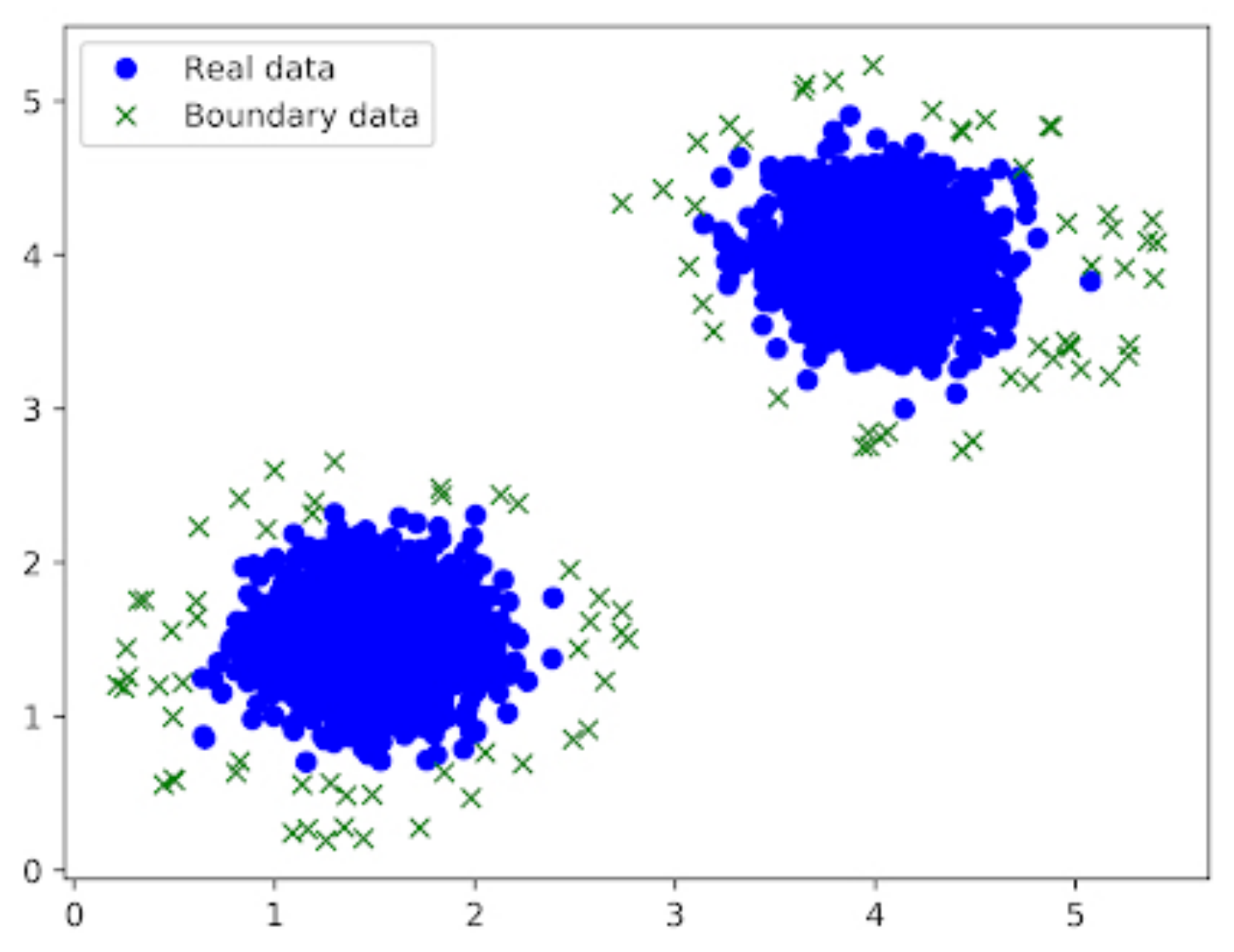}
    \captionof{figure}{OMASGAN Task 2 for synthetic data \\ where the blue points are $\textbf{x}$ and the green $B(\textbf{z})$.}
    \label{fig:3}
\end{wrapfigure}

\textbf{Evaluation of OMASGAN on synthetic data.}
We train the proposed f-divergence-based OMASGAN on synthetic data in Figure~\ref{fig:3}. The evaluation on synthetic data shows that OMASGAN generates abnormal OoD data on the boundary of the distribution support.
Figure~\ref{fig:3} shows the OMAS samples, i.e. the green $B(\textbf{z})$ points. The blue points are samples from the data distribution.
We show that OMASGAN eliminates mode collapse and works for multimodal data distributions with a support with \textit{disconnected components} \cite{31, 37, 30}.
Our scattering term in (3) and (5) preserves distance proportionality in $\mathscr{Z}$ and $\rchi$, eliminating mode collapse.
OMASGAN achieves dispersion using $s( B(\textbf{z}), \textbf{z} )$ in Section~\ref{sec:asdfsadfsdafsfsdfss}, not requiring the calculation of the center of mass of the data distribution.
OMASGAN obviates the sampling complexity problem and the parallel estimation of the mass and the boundary.
We continue the evaluation of OMASGAN on synthetic data, and we evaluate OMASGAN using histograms of anomaly scores. With retraining, we increase the AUROC metric and the Area Under the Precision-Recall Curve (AUPRC) from $0.91$ to $0.99$ and the F1 score, Precision, Recall, and Accuracy from $0.83$ to $0.98$ on a grid of equidistant points.

\begin{figure}[!bp]
\centering
\begin{minipage}[b]{.49\textwidth}
  \centering
  \includegraphics[width=1.\linewidth]{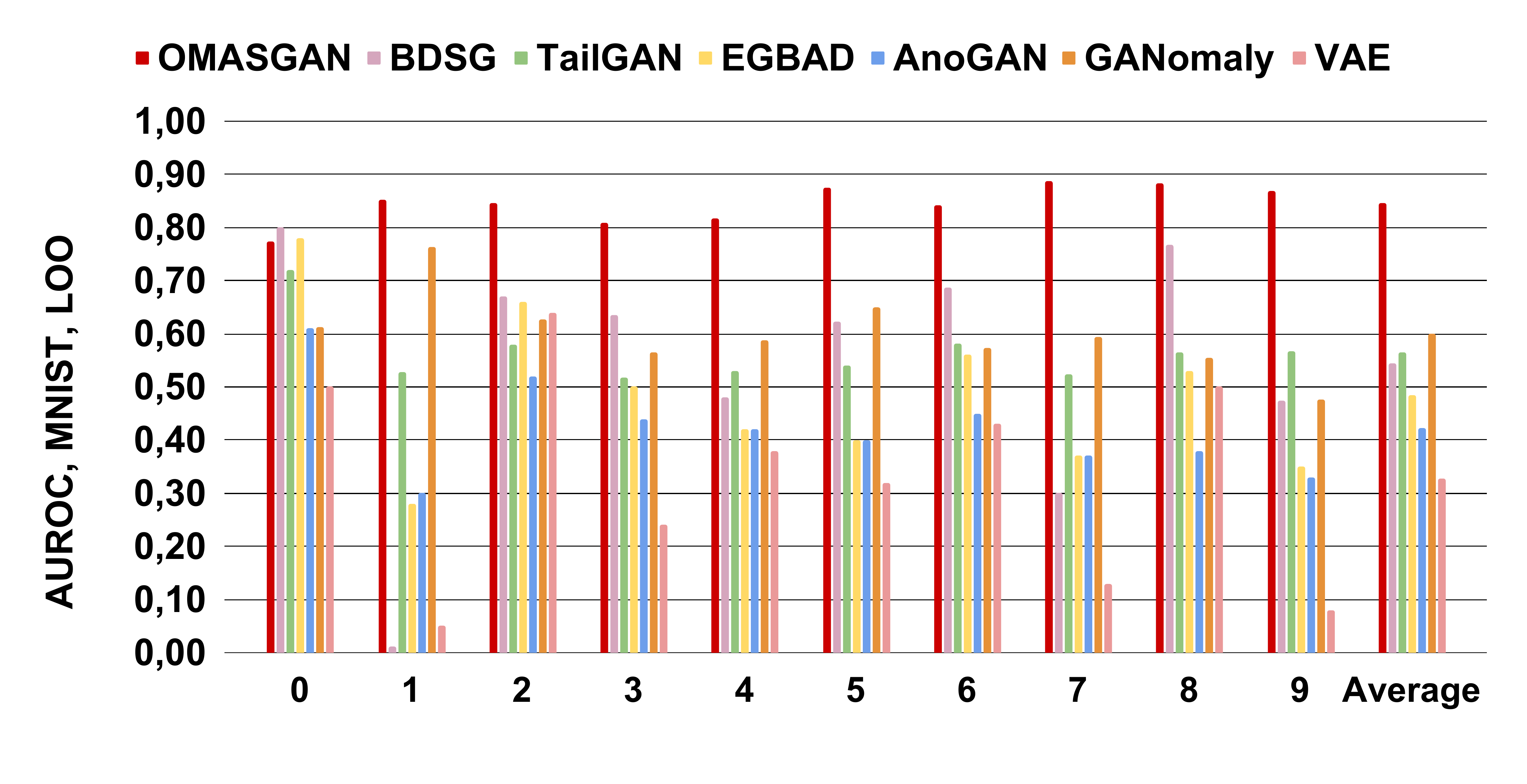}
  \captionof{figure}{Performance of OMASGAN on MNIST in AUROC compared to GAN and AE with LOO.}
  \label{fig:4}
\end{minipage}%
\hfill
\begin{minipage}[b]{.49\textwidth}
  \centering
  \includegraphics[width=1.\linewidth]{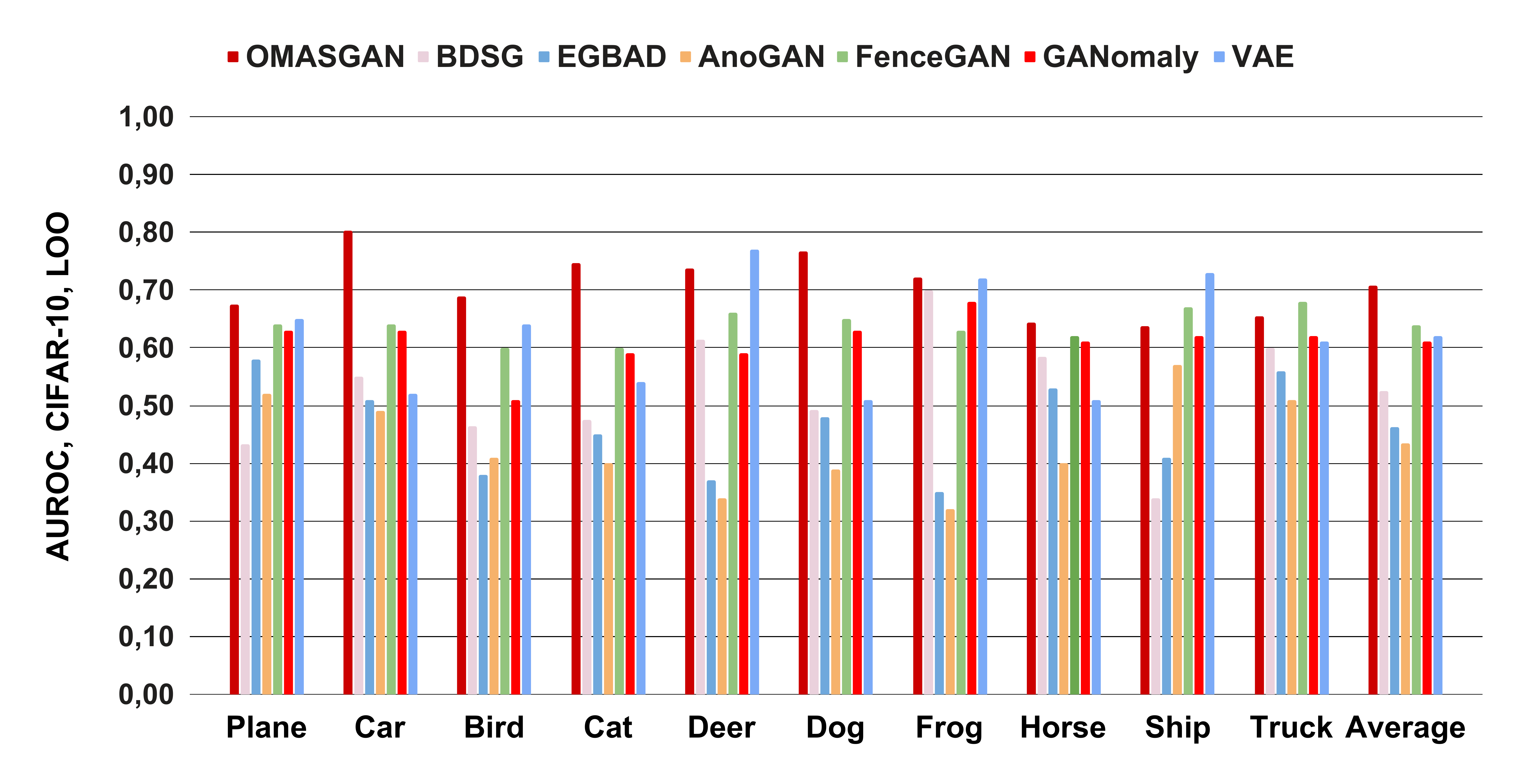}
  \captionof{figure}{Performance of OMASGAN in AUROC on CIFAR compared to GAN and AE with LOO.}
  \label{fig:5}
\end{minipage}
\end{figure}

\textbf{Evaluation of OMASGAN on MNIST.} \label{sec:jdsdfshfjgy66tdd} 
\textit{Setup:} We evaluate the proposed f-divergence-based OMASGAN model using the LOO methodology.
We train OMASGAN using the KLWGAN until convergence, approximately $500$ epochs.
We use CNN architectures for generating the distributions $G(\textbf{z})$, $B(\textbf{z})$, and $G'(\textbf{z})$.
We use $p_{\textbf{z}}=\text{N}_{128}(\textbf{0}, \textbf{1})$, $Q = 1024$, $N = 256$, $\mu = 0.2$, and $\nu = 0.25$ in Section~\ref{sec:asdfsadfsdafsfsdfss}.
We train OMASGAN and successfully generate $G(\textbf{z})$, $B(\textbf{z})$, and $G'(\textbf{z})$.
We evaluate OMASGAN and its discriminator $J(\textbf{x})$ using histograms of the anomaly scores for normal and abnormal samples.

\textit{LOO Evaluation in AUROC:}
Figure~\ref{fig:4} shows that on average and for all the MNIST digits, OMASGAN outperforms the GAN benchmarks EGBAD, AnoGAN, BDSG, and TailGAN.
Also, Figure~\ref{fig:4} shows that OMASGAN outperforms the AE-based models GANomaly and VAE.
We evaluate OMASGAN and compare it with GANomaly using the same inference conditions as those in OMASGAN, i.e. training set statistics rather than test set statistics for the batch normalization layers.
Figure~\ref{fig:4} shows that OMASGAN achieves on average an AUROC of $0.85$ on MNIST and outperforms the benchmarks by at least $0.24$ points in AUROC, by a percentage of $41\%$.
The proposed OMASGAN model is robust and achieves the lowest standard deviation (SD) averaged over all the MNIST digits, i.e. $0.036$, compared to EGBAD, AnoGAN, BDSG, TailGAN, GANomaly, and VAE. These AD benchmarks have SDs averaged over all the digits $0.153$, $0.093$, $0.24$, $0.059$, $0.074$, and $0.199$, respectively.

\textbf{Evaluation of OMASGAN on CIFAR-10.} \label{sec:sfgafg234234dfa}
\textit{Setup:} In \eqref{eq:dasfadfadsfa1xsx1x} and \eqref{eq:sdafasdf98sadf766sdafas} in Section~\ref{sec:hdsgahg32asf1d1}, we use $\alpha+\gamma=0.7$, $\beta=0.7$, and $\delta=0.5$, as in \cite{6, 51}.
\textit{LOO Evaluation in AUROC:}
Figure~\ref{fig:5} shows that the performance of OMASGAN using LOO is better than that of the GAN models AnoGAN, EGBAD, FenceGAN, and BDSG on average and for all tasks.
In Figure~\ref{fig:5}, on average and for almost all classes, the proposed OMASGAN outperforms the AE benchmarks GANomaly, VAE, ADAE, and AED.
According to Figure~\ref{fig:5}, OMASGAN outperforms the benchmarks in AUROC averaged over all classes. It is robust achieving the lowest SD, $0.056$, compared to the AD benchmarks. 
It outperforms the benchmarks on average over all tasks by at least $0.07$ AUROC points, by a percentage increase of at least $11\%$.
OMASGAN achieves on average an AUROC of $0.71$ on CIFAR-10 using LOO evaluation.
On the CIFAR-10 dataset, it achieves an improvement of approximately $7.5\%$ compared to benchmarks.

\begin{figure}[!tbp]
\centering
\begin{minipage}[b]{.49\textwidth}
  \centering
  \includegraphics[width=1.\linewidth]{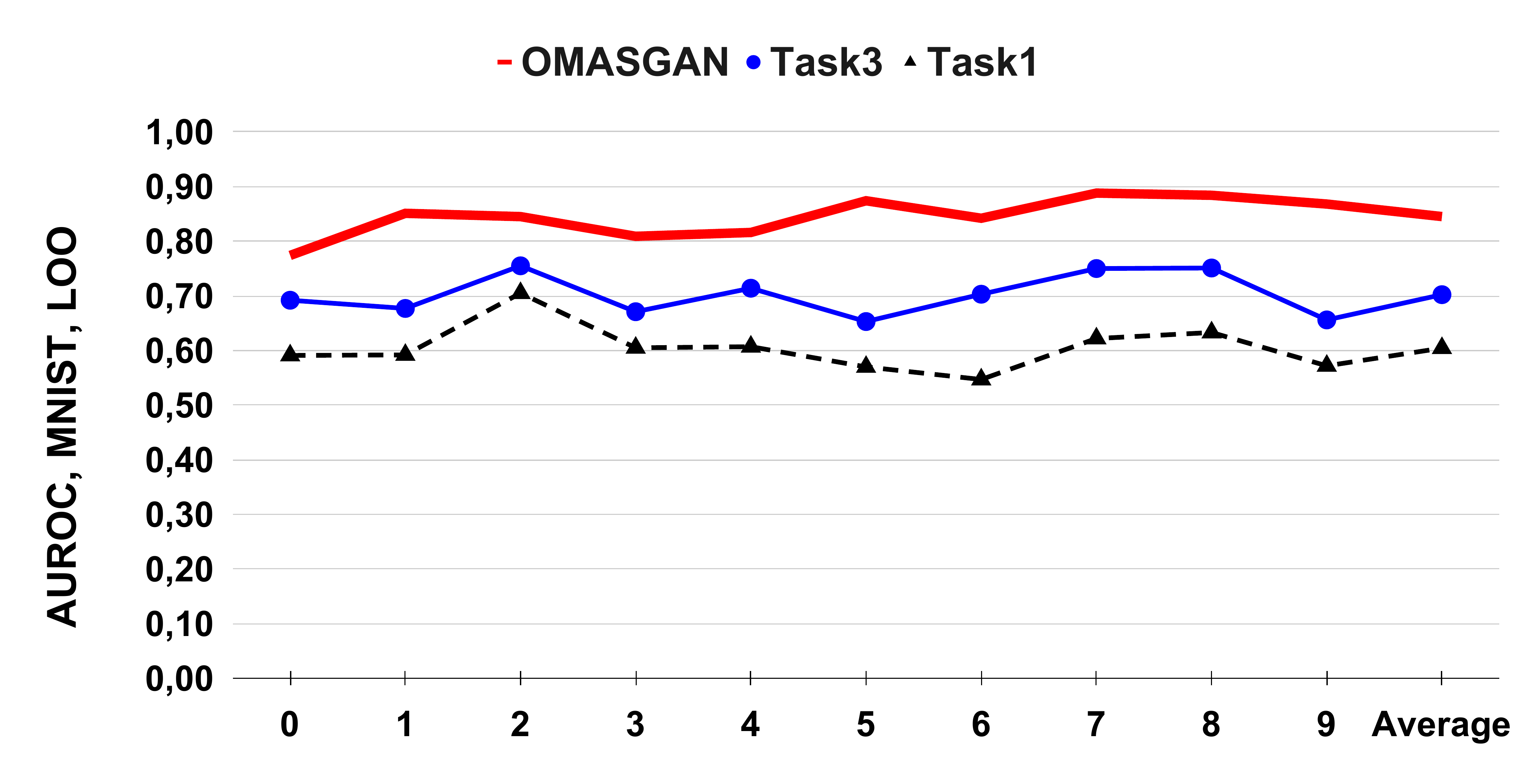}
  \captionof{figure}{Ablation study of OMASGAN in A-\\UROC on MNIST data using LOO evaluation.}
  \label{fig:6}
\end{minipage}%
\hfill
\begin{minipage}[b]{.49\textwidth}
  \centering
  \includegraphics[width=1.\linewidth]{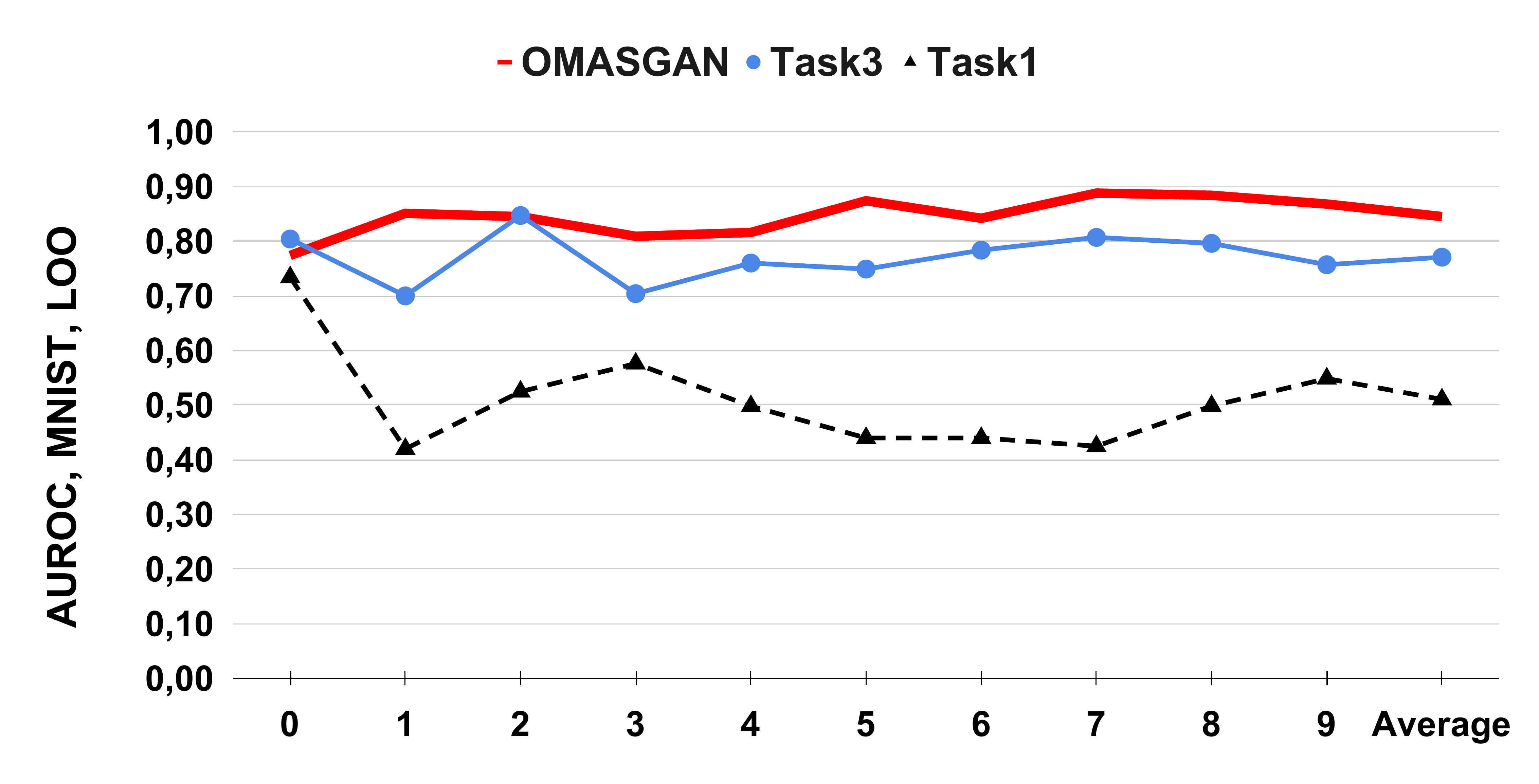}
  \captionof{figure}{Comparison of OMASGAN to the f-GAN-based OMASGAN using LOO on MNIST.}
  \label{fig:7}
\end{minipage}
\centering
\begin{minipage}[b]{.49\textwidth}
  \centering
  \includegraphics[width=1.\linewidth]{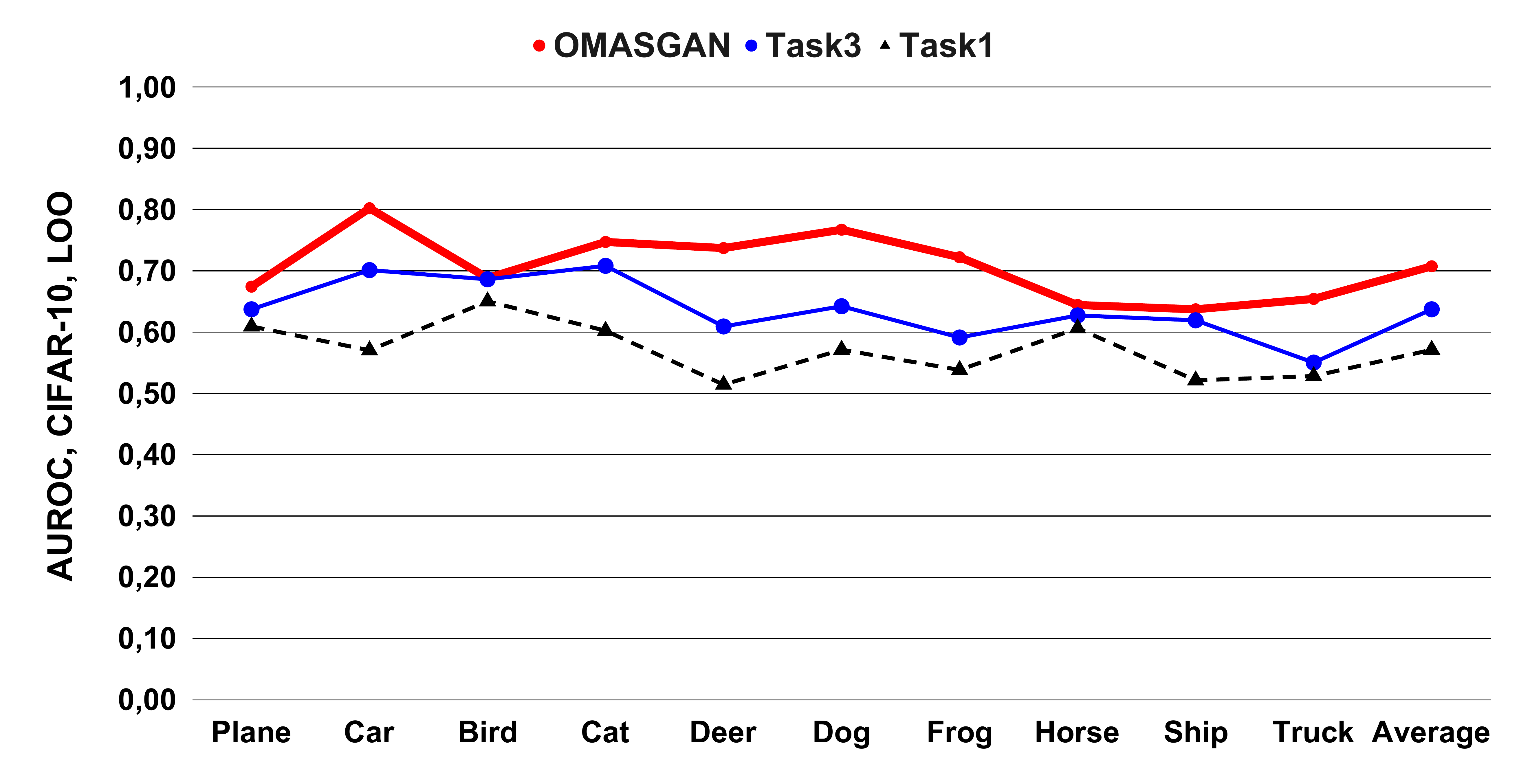}
  \captionof{figure}{Ablation study of OMASGAN in AUROC on CIFAR: Impact of the losses using LOO.}
  \label{fig:8}
\end{minipage}%
\hfill
\begin{minipage}[b]{.49\textwidth}
  \vspace{7.88pt}
  \centering
  \includegraphics[width=0.983\linewidth]{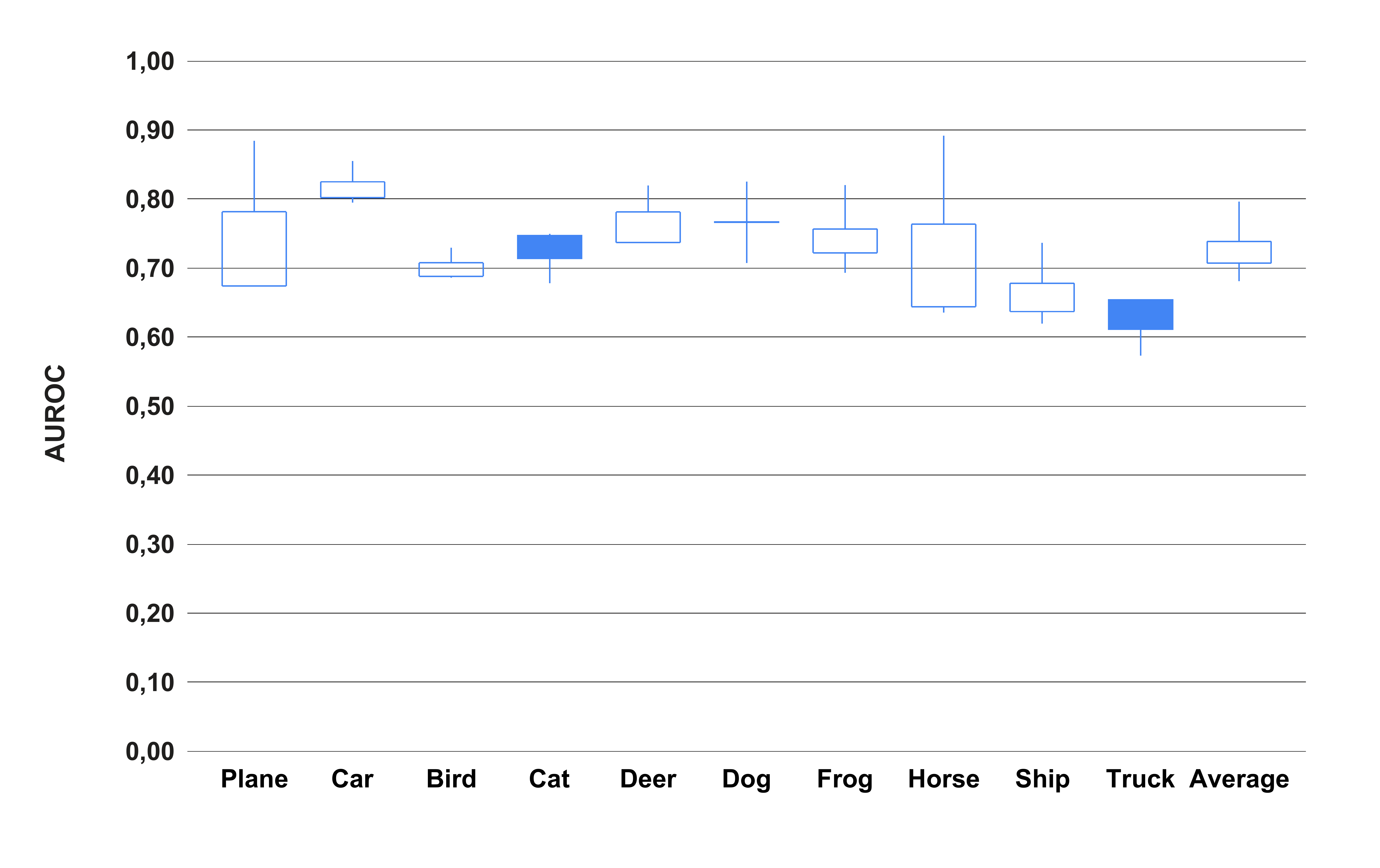}
  \captionof{figure}{Performance of OMASGAN using LOO on CIFAR-10 data for random seeds $0$, $1$, and $2$.}
  \label{fig:9}
\end{minipage}
\end{figure}

\textbf{Ablation study/analysis of OMASGAN.}
\textit{Benefit of model retraining by including negative boundary samples:} \textit{OMASGAN trained on MNIST:} Figure~\ref{fig:6} shows that, on average and for all the MNIST digits, OMASGAN improves the performance of the KLWGAN model implemented in Task 1 for OoD detection.
Comparing the training objective in Task 1 to the loss in Task 3 and to the final loss, OMASGAN improves the performance of the base model. 
The base model KLWGAN achieves an AUROC of $0.59$ averaged over all the MNIST digits, increasing to $0.71$ using Task 3, and then to $0.84$ using our final OMASGAN model. This is the contribution of our proposed Tasks 2 and 3.

\textbf{Effect of base model and chosen f-divergence.}
In Figure~\ref{fig:7}, we compare OMASGAN using the KLWGAN to the OMASGAN using the f-GAN on MNIST in AUROC \cite{44, 34}.
According to the ablation study, OMASGAN improves the performance of f-GAN by $0.26$ on average over all digits.
The base model f-GAN achieves on average an AUROC of $0.51$, which increases to $0.77$ because of OMASGAN Task 3.
Figure~\ref{fig:7} shows the benefit of the OMASGAN properties of \textit{active negative sampling and training}, learned negative data augmentation, boundary loss training, and discriminator anomaly score inference, presented in Table~\ref{tab:1}. We also observe robustness across the AD tasks.

\textbf{Improvement of OMASGAN compared to KLWGAN for AD on CIFAR-10.}
Figure~\ref{fig:8} presents an ablation study on the losses of OMASGAN on CIFAR-10 in AUROC using the LOO evaluation methodology.
We examine the impact of all our objective cost functions on the anomaly and OoD detection performance in Figure~\ref{fig:8}.
Our chosen base model, KLWGAN, yields an AUROC of $0.57$ averaged over all the LOO classes and this increases to $0.64$ using OMASGAN Task 3 and to $0.71$ using OMASGAN.
In reference to Table~\ref{tab:1}, Figure~\ref{fig:8} shows that the improvement of $0.14$ points in AUROC is the contribution of our \textit{retraining and negative data augmentation methodology}.
This is the benefit of the proposed active negative and boundary loss training in Section~\ref{sec:asdfsadfsdafsfsdfss}.
The average SD over all the LOO classes is $0.05$, $0.05$, and $0.06$ for Task1, Task3, and OMASGAN, respectively.

\textbf{Effect of selected inference mechanism.}
Following our discussion in Section~2, the anomaly score for a test sample $\textbf{x}^*$ is fD($G'$, $\delta_\textbf{x}^*$) and fD($G$, $\delta_\textbf{x}^*$) if the OMASGAN model is stopped at Task 3 and Task 1, respectively, in the ablation study in Figures~\ref{fig:6}-\ref{fig:8}. 
Comparing OMASGAN to Tasks 1 and 3, the performance \textit{gradually improves} and the use of $J(\textbf{x}^*)$, as presented in Section~\ref{sec:sfhghjsf7safgs}, is beneficial.

\textbf{Sensitivity to initialization.}
Figure~\ref{fig:9} shows the sensitivity of OMASGAN on CIFAR-10 to changes to the random seeds $0$, $1$, and $2$.
It shows the mean AUROC over all seeds per LOO class, the AUROC for seed $2$, and the mean AUROC $\pm$ SD.
The performance of OMASGAN yields a $0.05$ difference between seeds.
The mean SD over all classes and seeds is $0.06$ and this is \textit{lower} than that of \cite{37} which is $0.1$.
The mean SD per seed over all classes is $0.06$ for seed $2$, and $0.09$ for seeds $0$ and $1$.

\begin{wrapfigure}{r}{0.50\textwidth} 
    \vspace{-7pt}
    \centering
    \includegraphics[height=0.266\textwidth]{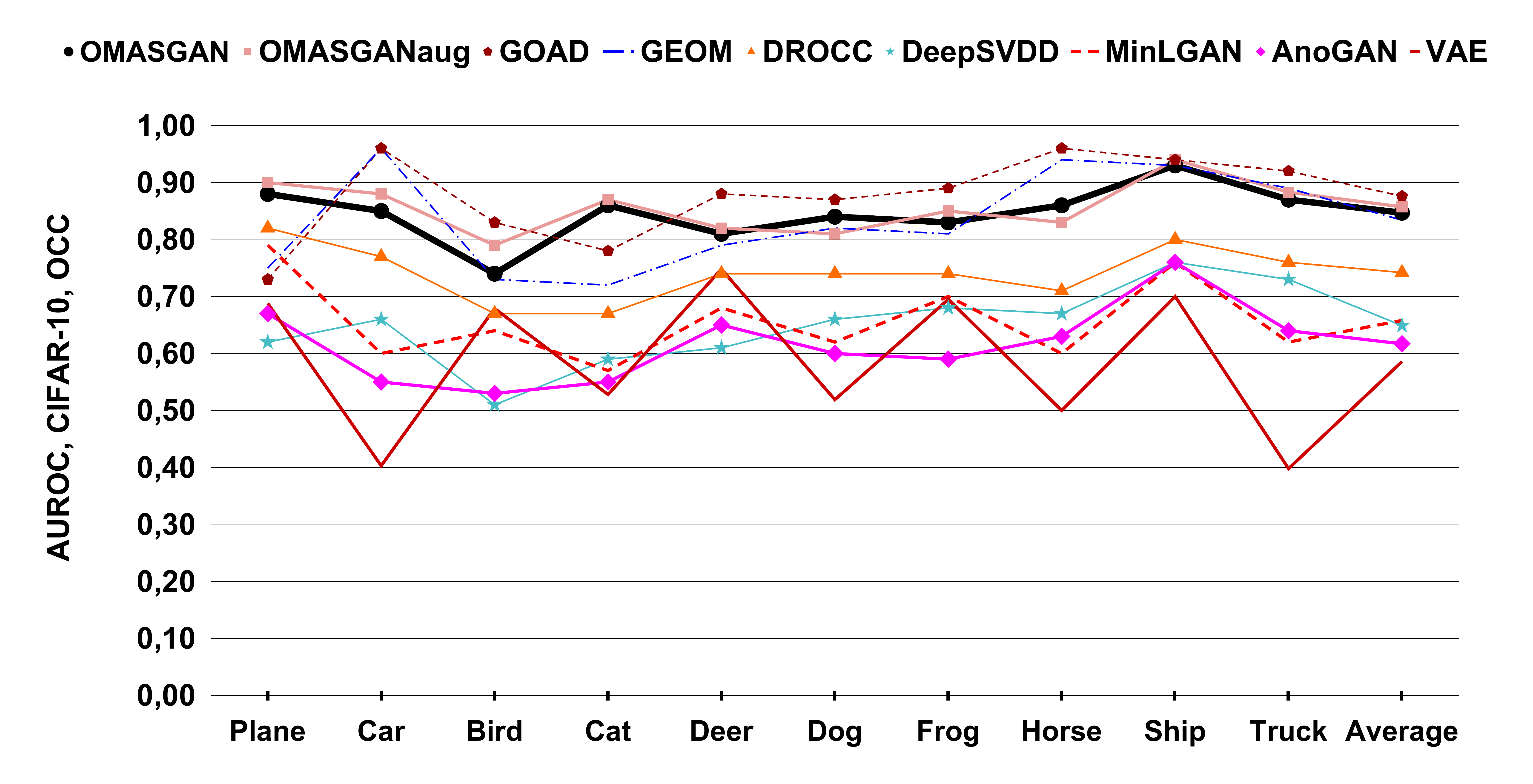}
    \captionof{figure}{Performance of our OMASGAN in \\
    AUROC on CIFAR-10 data using OCC evaluation.}
    \label{fig:10}
\end{wrapfigure}

\textbf{Evaluation of OMASGAN on CIFAR-10 using OCC.}
In Figure~\ref{fig:10}, we evaluate OMASGAN and compare it to GAN and AE models, and to GOAD and GEOM.
Our model achieves an average AUROC of $0.85$.
It outperforms the GAN models AnoGAN and MinLGAN, and the AE models VAE and DeepSVDD.
OMASGAN also outperforms the discriminator-based model DROCC. 
OMASGAN achieves robustness across the AD tasks.
OMASGANaug uses data augmentation comprising geometric image transformations, i.e. horizontal flipping and color augmentation, during training and slightly improves the AD performance of OMASGAN.
The performance of OMASGAN and OMASGANaug is comparable to that of the classification model GEOM.
GOAD uses data augmentation comprising geometric image transformations, e.g. flips and rotations, as well as affine transformations, and slightly outperforms OMASGANaug on average by approximately $2.3 \%$.
The classification-based models GEOM and GOAD use features to discriminate between transformations. 
On the contrary, OMASGAN does not use \textit{feature engineering}, human intervention, and manual processes. This is desirable and strengthens scalability and applicability.

\section{Discussion and Conclusion} \label{sec:safas23432dsfs4w5dsfs12232df}
We have proposed OMASGAN, a retraining methodology for AD with active negative sampling and training and self-supervised learning.
OMASGAN performs negative retraining by including the generated boundary which has the effect of ``pushing'' the distribution away from the OoD samples to improve the learning of the data distribution.
AD is a direct outcome of retraining which can be used in conjunction with existing models.
The proposed retraining methodology does not make any assumptions about the underlying data distribution, is not ad hoc, and is data- and architecture-agnostic.
OMASGAN performs learned data augmentation and AD using negative sampling.
Without requiring invertibility or likelihood, we generate OMAS samples and \textit{strong anomalies} leveraging any f-divergence in its variational representation.
OMASGAN does not use feature engineering and works for data distributions with a support with disconnected components without mode collapse.
We address the \textit{rarity of anomalies} using data only from the normal class.
We use the discriminator to compute distribution metrics as they permit the modelling of rarity, taking into account the weak property of f-divergences and the evolution of generative models from AEs to GANs.
The evaluation outcomes on MNIST and CIFAR-10, as well as on synthetic data, using the LOO methodology show that OMASGAN achieves state-of-the-art performance and outperforms the GAN- and AE-based benchmarks, as illustrated in Figures~\ref{fig:4} and \ref{fig:5}.
The ablation study in Figures~\ref{fig:6}-\ref{fig:8} shows that OMASGAN improves the base model for AD using LOO evaluation on MNIST and CIFAR-10. 
Using the AUROC, OMASGAN yields on average (i) an improvement of at least $0.24$ points on MNIST over the benchmarks, achieving values of $0.85$, and (ii) an improvement of at least $0.07$ points on CIFAR-10, achieving values of $0.71$ using the LOO methodology.
LOO evaluation is more realistic than OCC; in real-world scenarios, we have a large number of items that are not rare and we are interested in detecting the objects that are significantly different from these items.
Figure~\ref{fig:10} illustrates that OMASGAN outperforms the GAN and AE benchmarks MinLGAN, AnoGAN, VAE, DeepSVDD, and the discriminator-based DROCC on CIFAR-10 using OCC evaluation, and achieves an average AUROC of $0.85$.
OMASGAN also achieves robustness across the different AD tasks.

\vfill
\pagebreak

\vfill
\pagebreak

\appendix

\section*{Appendix and Supplementary Material}
The discussions, experiments, and evaluations in this Section are a continuation of the paper ``OMASGAN: Out-of-Distribution Minimum Anomaly Score GAN for Sample Generation on the Boundary''.

\vspace{4pt}
\section*{A. OMASGAN and Illustration of our Algorithm}
\vspace{-119pt}
\begin{figure}[!htbp]
    \centering
    \begin{center}
    \begin{minipage}[!htbp]{0.93\columnwidth}%
    \begin{center}
	\centering \includegraphics[width=0.89\columnwidth]{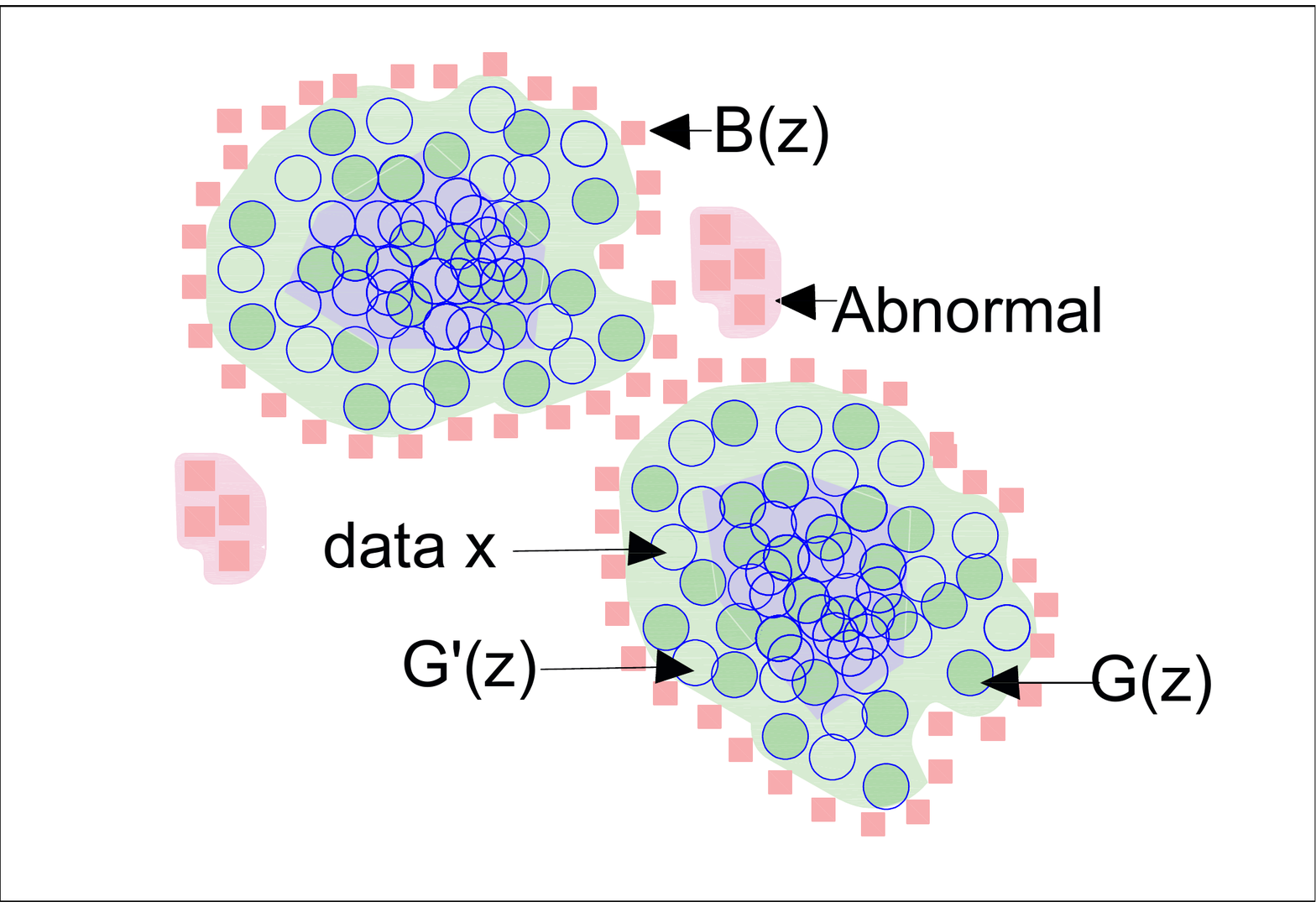}
    \end{center}
	\end{minipage}\hfill{}%
    \end{center}
	\vspace{-119pt}
	\caption{Illustration and pictorial representation of the proposed OMASGAN algorithm for AD.}
	\label{fig:tytytrewe21a}
\end{figure}

\vspace{4pt}
In this section, we present an illustration of the proposed OMASGAN algorithm using the notation and the implicit generative models introduced in Section~2 of the paper.

Figure~\ref{fig:tytytrewe21a} depicts the main idea of the proposed OMASGAN algorithm.
OMASGAN first generates minimum-anomaly-score OoD samples created by our negative sampling augmentation methodology and then performs active negative training for AD. OMASGAN performs model retraining and subsequently trains a discriminative model for AD using the generated samples on the boundary of the support of the data distribution.

\vfill

\pagebreak
\vspace{4pt}
\section*{B. Implementation of OMASGAN}
The proposed OMASGAN model is implemented in PyTorch: \href{https://github.com/Anonymous-Author-2021/OMASGAN}{https://github.com/Anonymous-Author-2021/OMASGAN}

Online:~\href{https://anonymous.4open.science/r/OMASGAN-5702/README.md}{https://anonymous.4open.science/r/OMASGAN-5702/README.md}

\vspace{4pt}
\section*{C. Images from OMASGAN Tasks 1 and 3}
\begin{figure}[!htbp]
    \begin{center}
	\begin{minipage}[!htbp]{0.77\columnwidth}%
		\centering
        \begin{center}
		\includegraphics[width=0.77\columnwidth]{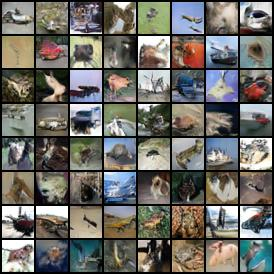}
		\centering
		\includegraphics[width=0.77\columnwidth]{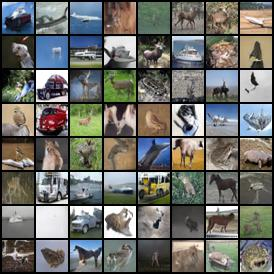}
        \end{center}
		\end{minipage}\hfill{}%
  	    \caption{Images from the normal class generated by the proposed OMASGAN model from Task 1 (upper) and Task 3 (lower), trained on CIFAR-10 data using the LOO evaluation methodology.}
    \end{center}
\end{figure}

\vfill
\pagebreak

\end{document}